\documentclass[10pt, a4paper]{article}

\usepackage{amssymb,amsfonts,bm}
\usepackage{makecell}

\usepackage{url}

\usepackage{breakurl}
\usepackage[breaklinks]{hyperref}
\usepackage{latexsym}
\usepackage{graphicx} 
\usepackage{microtype}
\usepackage{multirow}
\usepackage{multicol}
\usepackage{booktabs}
\usepackage{lrec-coling2024} 

\title{Guided Distant Supervision for Multilingual Relation Extraction Data: Adapting to a New Language}

\name{Alistair Plum$^\diamondsuit$, Tharindu Ranasinghe$^\spadesuit$, Christoph Purschke$^\diamondsuit$} 

\address{$^\diamondsuit$University of Luxembourg, Esch-sur-Alzette, Luxembourg\\
        $^\spadesuit$Aston University, Birmingham, UK\\
        \{\texttt{alistair.plum,christoph.purschke}\}\texttt{@uni.lu}\\
        \texttt{t.ranasinghe@aston.ac.uk}}

\abstract{
Relation extraction is essential for extracting and understanding biographical information in the context of digital humanities and related subjects. There is a growing interest in the community to build datasets capable of training machine learning models to extract relationships. However, annotating such datasets can be expensive and time-consuming, in addition to being limited to English. This paper applies \textit{guided distant supervision} to create a large biographical relationship extraction dataset for German. Our dataset, composed of more than 80,000 instances for nine relationship types, is the largest biographical German relationship extraction dataset. We also create a manually annotated dataset with 2000 instances to evaluate the models and release it together with the dataset compiled using guided distant supervision. We train several state-of-the-art machine learning models on the automatically created dataset and release them as well. Furthermore, we experiment with multilingual and cross-lingual experiments that could benefit many low-resource languages. 
 \\ \newline \Keywords{Relation Extraction, Distant Supervision, Language Resources} }

\begin{document}

\maketitleabstract

\section{Introduction}
Biographical information extraction (IE) is the process of extracting factual information about people from unstructured and semi-structured data and storing it in a structured format \cite{khirbat-etal-2016-n}. The information can contain essential facts or events related to the lives of well-known and lesser-known individuals. The extracted structured information has many applications ranging from humanities to computer science.

Two essential components of an IE system are entity extraction and relation extraction (RE). In classical IE systems, these components appeared separately and sequentially \cite{finkel-etal-2005-incorporating, zelenko2003kernel,chan-roth-2011-exploiting}. However, such IE systems ignore the interactions between the two components \cite{shang2022onerel} and are susceptible to the problem of error propagation \cite{li-ji-2014-incremental}. Therefore, recent studies focus on building joint models to obtain entities together with their relations through a unified architecture \cite{yamada-etal-2020-luke,8983370,zhang-etal-2017-position,yu2020joint}. These systems extract pairs of entities and their relationship. 

Several machine learning models, including the latest deep learning methods \cite{Nayak_Ng_2020,joshi-etal-2020-spanbert,zeng-etal-2014-relation} have been proposed to jointly extract entities and their relationships from text, which we describe more in Section \ref{sec:related}. These machine learning models follow a supervised approach, where the models require a large amount of high-quality data for training. Following this, several manually annotated datasets have been released for biographical RE \cite{hoffmann-etal-2011-knowledge,zhang-etal-2017-position}. However, manual annotation is expensive and time-consuming \cite{surdeanu-etal-2012-multi}. Therefore, manually annotated datasets are often restricted to English, and specific domains \cite{10.1609/aaai.v33i01.33017418}.

\textit{Distant supervision} for RE has been proposed to create a large amount of automatically generated labels \cite{mintz-etal-2009-distant}, which is based on the assumption that any sentences containing two related entities will express this relation. Although distant supervision is popular and the majority of recent multilingual datasets are based on this paradigm \cite{seganti-etal-2021-multilingual}, it has major flaws \cite{hoffmann-etal-2011-knowledge} due to its idealistic hypothesis. 


To overcome this, \textit{guided distant supervision} (GDS) was proposed recently by \citet{plum2022} with the aim of ensuring correct labels for the relations identified by distant supervision by using external resources, such as Pantheon \cite{Yu2016}, and Wikidata \cite{wikidata2014}. As the initial work on GDS was originally conceptualised for English, we propose a novel large dataset for German biographical RE by adapting GDS and overcoming several challenges in the methodology.
Finally, we address the possibility of adapting GDS to further low-resource languages by experimenting with cross-lingual RE. Our approach significantly reduces the burden on the manual annotation process while having more accurate sentences than traditional distant supervision.  

The \textbf{main contributions} of this paper are:

\begin{enumerate}
    \item We introduce the largest German dataset for biographical RE built using GDS with ten relationship categories. We also produce a manually annotated subset that can be used for evaluation.

    \item We evaluate several machine learning models to perform biographical RE for German, based on state-of-the-art transformer models such as BERT \cite{devlin2019bert} and XLM-RoBERTa \cite{conneau-etal-2020-unsupervised}.
    
    \item We explore cross-lingual transfer learning between two datasets created using GDS. We show that large datasets created using GDS can be used to perform RE using cross-lingual transfer learning.

    \item We provide important resources to the community: the dataset, the code, and the pre-trained models are made available\footnote{\url{https://huggingface.co/datasets/plumaj/biographical}} to everyone interested in working on biographical RE using the same methodology.
\end{enumerate}


\section{Related Work}
\label{sec:related}
Due to its wide range of applications, biographical IE from online documents is a popular research area in the NLP community. As mentioned, there are two primary components in a traditional IE system; entity extraction and relation extraction \cite{finkel-etal-2005-incorporating}. \citet{mintz-etal-2009-distant} and \citet{10.1007/978-3-642-15939-8_10} use pipeline approaches where a named entity recognition (NER) system is used to identify the entities in a sentence, followed by a classifier to determine the relation between them. However, due to the complete separation of entity extraction and relation extraction, these models miss the interaction between multiple relation tuples present in a sentence \cite{Nayak_Ng_2020} and can also propagate errors from one component to the other \cite{ZHENG201759}. Due to this limitation, researchers suggested joint entity and relationship extraction for IE systems. These approaches extract triplets that contain two entities and the relationship between them \cite{Nayak_Ng_2020}. 

Recently, several neural network-based models \cite{katiyar-cardie-2016-investigating, miwa-bansal-2016-end} were proposed to extract entities and relations from a sentence jointly. The current state-of-the-art in RE, also used for this research, is based on neural transformers \cite{baldini-soares-etal-2019-matching}. These transformer models are trained using a language modelling task such as masked language modelling or next sentence prediction and then have been used to perform RE as a downstream NLP task. Results on recent RE datasets indicate that transformers outperform the previous architectures based on RNNs and CNNs \cite{8983370,baldini-soares-etal-2019-matching}. 

The machine learning models previously referred to follow a supervised paradigm, where a dataset is required to train. Therefore, the NLP community has a growing interest in producing datasets capable of training machine learning models to perform RE. Several datasets in this area, such as NYT24 \cite{hoffmann-etal-2011-knowledge} and TACRED \cite{zhang-etal-2017-position}, have been released for this purpose. As we mentioned before, annotating data manually for RE is time-consuming and expensive, limiting manually annotated datasets largely to English. 

As a solution, \citet{mintz-etal-2009-distant} proposed distant supervision to automatically label RE datasets. Here, the assumption is made that if there is a relation between entities, then every sentence containing these two entities may also express that relation. Several multilingual datasets for biographical RE have been released following the distant supervision paradigm \cite{seganti-etal-2021-multilingual}. However, the base hypothesis of distant supervision that all instances containing the same entity pairs express the same relation is not always correct due to the existence of \textit{multi relations} between entities. Various solutions such as multi-instance learning \cite{hoffmann-etal-2011-knowledge} and sentence-level attention \cite{lin-etal-2016-neural} have been proposed to overcome this, but these approaches only select the instance with the highest probability of being a valid candidate, so a large amount of rich information can be lost. Recently, \citet{plum2022} proposed GDS to create biographical RE datasets automatically. Because the study was limited to English, and the use of external resources makes it difficult to port it to different languages, we aim to address those gaps in this research by adapting GDS to create a German biographical RE dataset.

\section{Guided Distant Supervision}
\label{sec:method}
As highlighted in the previous section, adapting data compilation methods to other languages can be difficult due to resource and labour requirements. The approach presented here shows that these limitations can be overcome using resources that are multilingual by design (although these do not necessarily guarantee sentence-level alignment), and combining these with an automatic labelling approach. The steps are split into data source selection (Section \ref{sec:datsources}), and automatic labelling (Section \ref{sec:label}).


\subsection{Data Sources}\label{sec:datsources}
The approach for our data compilation method requires a source of textual data and a structural counterpart, much like distant supervision \cite{mintz-etal-2009-distant}. Therefore, we use Wikipedia, Wikidata, and Pantheon\footnote{\url{https://pantheon.world}} \cite{Yu2016}. Each data source provides different information for the automatic compilation process.

Pantheon serves as our main source of structured data, as well as an indicative list of entities to target. 
The creators state that Pantheon is \emph{"focused on biographies with a presence in 15 different languages in Wikipedia"} and consists of roughly 85,000 partly manually curated entries. This allows us to work with an indicative list of entities, with links to Wikidata and Wikipedia, and basic meta (birthdate, birthplace, occupation, etc.).

In addition to Pantheon, we make use of Wikidata, which contains structured information from Wikipedia. We mainly use Wikidata to enhance some of the information about the relevant entities that is not available in Wikipedia, and retrieve certain parts of information in German, such as occupation and place of birth, as these are only available in English in Pantheon. Relations that contain occupations for instance, can be difficult to translate automatically, as for instance, German has two genders for such nouns, which could only be inferred from context information in English. Locations, on the other hand, may be complex since they can be ambiguous, and therefore rely on context information to determine the correct translation. 

Wikipedia serves as our source of textual data. As a free online encyclopedia, it contains large amounts of information about people, which suits our aim of targeting biographical data. In addition, our indicative list of entities, taken from Pantheon, is structured around Wikipedia. Processing the textual data requires a number of steps in itself, and we follow a previously established workflow \cite{plum2019}. In summary, we extract all relevant articles from a Wikipedia dump using article IDs. Since we work with German articles, we map the Wikipedia IDs from Pantheon (provided for English articles only) to the German article IDs using a database dump of Wikipedia, which contains this information. After matching the relevant articles, we extract the articles using the wikiextractor\footnote{\url{https://github.com/attardi/wikiextractor}} package for Python, which converts from XML to plain text.

\subsection{Automatic Labelling}\label{sec:label}
We automatically label sentences taken from Wikipedia by processing the previously selected articles using spaCy\footnote{\url{https://spacy.io}}, running NER to tag entities, and then matching the entities to the structured data we have available. For this, we use the non-neural spaCy model for German since the latest neural model does not support NER. For the previously mentioned English counterpart to this dataset, the neural spaCy model was used. Therefore, we have worse performance here and have to work with far fewer supported entity types: location, person, organisation and miscellaneous. 

Below are two examples from the dataset (with English translations).

\vspace{1em}
\begin{itemize}
    \item[\textbf{DE:}] Im Alter von fast 77 Jahren starb <e1>\textbf{Lorenzo Ghiberti}</e1> am <e2>\textbf{1 Dezember 1455}</e2> in Florenz. -> \textit{\textbf{birthdate}}
    \item[\textbf{EN:}] At the age of almost 77 <e1>\textbf{Lorenzo Ghiberti}</e1> died on <e2>\textbf{1 Dezember 1455}</e2> in Florence. -> \textit{\textbf{birthdate}}
\end{itemize}
\vspace{0.5em}
\begin{itemize}
    \item[\textbf{DE:}] <e1>\textbf{Menger}</e1> lernte bei Hans Hahn und promovierte 1924 an der <e2>\textbf{Universität Wien}</e2>. -> \textit{\textbf{educated}}
    \item[\textbf{EN:}] <e1>\textbf{Menger}</e1> studied with Hans Hahn and received his doctorate from the <e2>\textbf{University of Vienna}</e2> in 1924. -> \textit{\textbf{educated}}
\end{itemize}
\vspace{1em}

For each article, we check the main article entity and then whether other matched entities are part of the structured information that we have available from our data sources. If an entity is matched to any information string, the sentence is labelled according to type of information. For all relations, we only accept the first occurrence of a match, as manual verification of sentences showed that accepting more than the first occurrence lead to many more incorrectly labelled sentences. This is due to the underlying assumption that the selected relations would appear towards the beginning of a Wikipedia article due to their importance.



As the NER model does not match dates or occupations, we use a loose regular expression to match dates. For the occupations, we match against a list of occupations gathered from the meta information from Pantheon. As the information in Pantheon is in English, we translated the list using deepL\footnote{\url{https://deepl.com}}, although since the list is quite short (around 100 items), we made some manual revisions where MT was not accurate, as well as making sure we have both masculine and feminine versions of each occupation, since German has distinct forms for these. To match the correct location names, we translated all places of birth and death using the \textit{alternate names} table provided by GeoNames. The names were matched via the alternate names table, and double checked through partial coordinate matching (first four digits). 

For matching the named entities to the structured Pantheon and Wikidata information, we allow list matching instead of string matching used in \citet{plum2022}. This was due to the fact that we wanted both the English and German versions of each relation to match since we found that the German counterpart of some relations (mainly the location names) was not always used.

We compiled two versions of the German dataset. While the normal method is performed as described above, the skip method involves skipping the first sentence of a Wikipedia article. The aim of this method is to find more diverse sentences since, many times, the first sentence of Wikipedia contains much of the desired information. As the first sentence has a very standardised structure (see below), it stands to reason that having many sentences of this kind could overfit a neural model unnecessarily, making it less precise with sentence types of a different structure.

\vspace{1em}
\begin{itemize}
    \item[\textbf{DE:}] Bernard Tomic (*21 Oktober 1992 in Stuttgart, Deutschland) ist ein australischer Tennisspieler.
    \item[\textbf{EN:}] Bernard Tomic (*21 October 1992 in Stuttgart, Germany) is an Australian tennis player.
\end{itemize}
\vspace{1em}

Three sets of the English Biographical dataset were compiled by \citet{plum2022} using different processing methods, the normal method as described here, one replacing all coreference mentions and one skipping the first sentence of an article. As spaCy does not support coreference resolution for German at the time of writing this paper, we only compiled two versions for the German dataset. 

Table \ref{tab:relations_count} shows the statistics for each relationship type in the German dataset, which we compare with the English dataset \cite{plum2022}. Overall, we found that for the German set, we were only able to gather a much smaller number of relations. This can be due to the fact that the performance of the German model for spaCy is not as precise as the English model. Furthermore, in the case of the \textit{skip} set we found a great deal fewer relations. We conjecture that the cause for this is the fact that the German articles may be much shorter in some places, therefore perhaps not containing all the desired relations. In turn, the German articles also feature more complex sentences in other places, therefore making it more difficult for the automatic labelling approach to match relations. This happened even though one of the main aims of using the Pantheon dataset was to ensure an even number of articles across languages.

\begin{table}[t!]
\centering
\begin{tabular}{|l|c|c|c|}
\hline
\textbf{Relation} & \textbf{EN} & \textbf{DE normal} & \textbf{DE skip} \\ \hline
birthdate      & 51,524           & 8,777           & 770             \\
birthplace     & 50,226           & 12,833          & 5,816           \\
child          & 2,209            & 718             & 701             \\
deathdate      & 17,197           & 922             & 454             \\
deathplace     & 18,944           & 4,059           & 3,263           \\
educatedAt     & 5,639            & 610             & 607             \\
occupation     & 18,114           & 10,861          & 4,836           \\
other          & 173,969          & 39,782          & 20,469          \\
parent         & 6,352            & 3,704           & 3,565           \\
sibling        & 2,083            & 917             & 890             \\ \hline
\textbf{Total} & \textbf{346,257} & \textbf{83,183} & \textbf{41,380} \\ \hline
\end{tabular}
\caption{Number of Relations in English \protect \cite{plum2022} vs. German Sets}
\label{tab:relations_count}
\end{table}

\subsection{Evaluation Data}\label{sec:evaldata}
To obtain a gold standard set of annotations, we separated 2000 sentences in total, with 100 sentences per relation from both processing methods, which were then manually annotated. We use this manual annotation to also obtain an insight into the performance of the automatic compilation method. Table \ref{tab:gold_eval} shows the evaluation results of the automatic compilation method. 

We used two native German speakers for the annotation. The guidelines were that each sentence should either explicitly or implicitly convey the relation, and no prior knowledge should have influence over the decision. While Cohen's Kappa for the inter-annotator agreement was calculated 0.92, which indicates a very high agreement between our annotators, cases of disagreement were discussed and assigned a label accordingly.

\begin{table}[t!]
\centering
    \begin{tabular}{|l|ccc|c|}
        \hline
        \textbf{Relation} & \textbf{P} & \textbf{R} & \textbf{F1} & \textbf{Supp.} \\ \hline
        birthdate & .98 & 1.0 & .99 & 196 \\
        birthplace & .69 & .83 & .76 & 167 \\
        deathdate & .92 & 1.0 & .96 & 184 \\
        deathplace & .20 & 1.0 & .33 & 39 \\
        educated & .92 & .99 & .95 & 184 \\
        occupation & .90 & 1.0 & .94 & 179 \\
        parent & .84 & .95 & .89 & 178 \\
        sibling & .65 & .99 & .78 & 130 \\
        child & .69 & .99 & .81 & 139 \\
        other & .94 & .31 & .47 & 604 \\ \hline
        \textbf{macro} & .77 & .91 & .79 & 2000  \\ \hline
    \end{tabular}
\caption{Evaluation of automatic labels against gold standard data.}
\label{tab:gold_eval}
\end{table}

\section{Neural Models}
\label{sec:models}
Using the described datasets for English and German, we trained a number of different neural models to examine relation extraction performance. All the neural models used are based on the neural transformers \cite{devlin2019bert}, and we make use of an architecture first shown by \citet{baldini-soares-etal-2019-matching}. Transformer models continue to show state-of-the-art results across many NLP objectives such as text classification \cite{ranasinghe-zampieri-2020-multilingual}, NER \cite{jia-etal-2020-entity}, question answering \cite{yang-etal-2019-end-end} and RE \cite{yamada-etal-2020-luke,joshi-etal-2020-spanbert,10.1145/3357384.3358119,alt2019improving}.

The model input is in sentence form, including markers for entity one \textit{[E1]} and entity two \textit{[E2]} positions. At these entity positions, the hidden states of the transformer are concatenated and used as the final output representation of a given relation. The last step involves stacking a linear classifier over the output representation. 


We fine-tune all the parameters from the transformer as well as the linear classifier jointly by maximising the log probability of the correct label. For all the experiments, we optimised parameters (with AdamW) using a learning rate of $7e-5$, a maximum sequence length of $512$, and a batch size of $32$ samples. The models were trained using a $24$ GB RTX 3090 GPU over five epochs. 
As stated previously, we use a number of pre-trained transformer models: \textit{bert-base-uncased} \cite{devlin2019bert}, \textit{bert-base-multilingual-cased} \cite{devlin2019bert}, \textit{bert-base-german-cased} \cite{chan-etal-2020-germans} and \textit{xlm-roberta-base} \cite{conneau-etal-2020-unsupervised}. All models were accessed via HuggingFace.

\section{Evaluation}
\label{sec:eval}
In this section, we present the evaluation results of the various neural models trained using different dataset combinations. The aim of this evaluation was to find out where the strengths and weaknesses of each data/model combination are, in order to verify whether the approach for data compilation can be successful.

\subsection{Baseline Results}\label{sec:multi:eval:base}
In order to provide a baseline result for this task, and to test the performance of machine translation (MT) as a potential alternative to GDS, we applied DeepL MT to the German evaluation set sentences to obtain English sentences. We then classified the English sentences using the pre-trained BERT model for English, which is fine-tuned on the English Biographical normal set \cite{plum2022}. The results of this classification are shown in Table \ref{tab:multi_baseline}. Overall, the results show that this approach performs well, similar to results reported in \cite{plum2022}, and highlight the performance of the original English Biographical dataset.

\begin{table}[t!]
	\centering
	\begin{tabular}{|l|c|c|c|}
		\hline
		\textbf{Relation} & \textbf{P} & \textbf{R} & \textbf{F1} \\ \hline
		birthdate         & .97        & .99        & .98         \\
		birthplace        & .79        & .91        & .85         \\
		deathdate         & .95        & .91        & .93         \\
		deathplace        & .30        & .89        & .45         \\
		educated          & .96        & .77        & .85         \\
		occupation        & .88        & .87        & .88         \\
		parent            & .75        & .92        & .83         \\
		sibling           & .92        & .84        & .88         \\
		child             & .86        & .67        & .75         \\
		other             & .73        & .66        & .69         \\ \hline
		macro avg.        & .81        & .84        & .81         \\ \hline
	\end{tabular}
	\caption[Evaluation of multilingual baseline approach]{Baseline results using MT with English model approach.}
	\label{tab:multi_baseline}
\end{table}

\subsection{Monolingual Learning}
We trained the transformer models on the German \textit{guided distantly supervised} dataset and evaluated using the German evaluation data. We used \textit{bert-base-multilingual-cased} \cite{devlin2019bert}, \textit{bert-base-german-cased} \cite{chan-etal-2020-germans} and \textit{xlm-roberta-base} \cite{conneau-etal-2020-unsupervised}. Table \ref{tab:monolingual_learn} shows the precision, recall and f1 score metrics for each relation, as well as overall macro and weighted scores. The XLM-R outperforms the other transformer models across many of the relationship types, and in terms of overall f1 score. Furthermore, we compare our results to the best results achieved for the English GDS dataset \cite{plum2022}. 
We observe that despite having less training data in the German dataset, the results are comparable between English and German. 

The results of the standard English BERT model trained on the English dataset and the German Bert model trained on the German dataset, both models evaluated using their respective Gold sets, are shown in Table \ref{tab:monolingual_learn}. While it is clear that overall the English model performs better, we find that in numerous instances, the German model achieves better recall (which is also evident in the total macro recall). We also see that \textit{occupation} is classified more precisely by the German model. 

\begin{table*}[ht!]
\centering
    \begin{tabular}{|l|ccc|ccc|ccc|ccc|}
    \hline
    \textit{Monolingual} & \multicolumn{3}{c|}{\textbf{GBERT}} & \multicolumn{3}{c|}{\textbf{mBERT}} & \multicolumn{3}{c|}{\textbf{XLM-R}} & \multicolumn{3}{c|}{BERT-EN}\\
    \textbf{Relation}  & \textbf{P} & \textbf{R} & \textbf{F1} & \textbf{P} & \textbf{R} & \textbf{F1} & \textbf{P} & \textbf{R} & \textbf{F1} & \textbf{P} & \textbf{R} & \textbf{F1} \\ \hline
    		birthdate  & .93 & .96 & \textbf{.95} & .93 & .96 & .94 & .94 & .96 & \textbf{.95} & 1.0 & 1.0 & 1.0 \\
    		birthplace     & .56 & .88 & .67 & .52 & .80 & .62 & .64 & .85 & \textbf{.78} & .91 & .91 & .91 \\
    		deathdate  & .92 & .97 & .94 & .91 & .96 & .93 & .94 & .95 & \textbf{.95} & .96 & .97 & .97  \\
    		deathplace     & .15 & .98 & .24 & .15 & .98 & .25 & .20 & .86 & \textbf{.30} & .31 & .73 & .41 \\
    		educated & .91 & .88 & \textbf{.89} & .92 & .79 & .85 & .93 & .81 & .86 & .94 & .83 & .88 \\
    		occupation & .89 & 1.0 & .94 & .89 & .98 & .93 & .92 & 1.0 & \textbf{.95} & .70 & 1.0 & .82 \\
    		parent  & .60 & .79 & .66 & .58 & .80 & .65 & .70 & .70 & \textbf{.75} & .81 & .74 & .78 \\
    		sibling & .64 & .79 & .70 & .63 & .87 & .72 & .72 & .75 & \textbf{.77} & .86 & .56 & .67 \\
    		child   & .61 & .54 & .56 & .67 & .54 & .60 & .87 & .56 & \textbf{.63} & .93 & .52 & .61 \\ 
    		other      & .75 & .49 & .59 & .76 & .49 & .60 & .79 & .76 & \textbf{.76} & .70 & .66 & .68 \\ \hline
    \textbf{macro}      & .69 & .84 & .72 & .69 & .83 & .71 & .75 & .79 & \textbf{.75} & .82 & .80 & .78 \\ \hline
    \end{tabular}
\caption{GBERT, mBERT and XLM-R trained on German GDS dataset and evaluated on German gold dataset. BERT-EN column shows best results for English dataset in \protect \citet{plum2022}.}
\label{tab:monolingual_learn}
\end{table*}

\subsection{Cross-lingual learning}
GDS relies on external data sources, such as Pantheon and Wikidata, which might not be available in low-resource languages. Therefore, in this section, we explore how transformer models trained with a GDS dataset perform with cross-lingual learning. We trained the \textit{bert-base-multilingual-cased} and \textit{xlm-roberta-base} \cite{conneau-etal-2020-unsupervised} transformer models on the English GDS dataset \cite{plum2022} and evaluated using the German evaluation data. Table \ref{tab:zero_learn} shows the precision, recall and f1 score for each relation, and overall macro.

\begin{table}[t!]
\centering
    \begin{tabular}{|l|ccc|ccc|}
        \hline
        \textit{Crosslingual} & \multicolumn{3}{c|}{\textbf{mBERT}} & \multicolumn{3}{c|}{\textbf{XLM-R}} \\
        \textbf{Relation} & \textbf{P} & \textbf{R} & \textbf{F1} & \textbf{P} & \textbf{R} & \textbf{F1} \\ \hline
        birthdate  & .93 & .93 & .93          & .95 & .94 & \textbf{.94} \\
        birthplace     & .63 & .75 & .68          & .65 & .73 & \textbf{.70} \\
        deathdate  & .93 & .86 & \textbf{.89} & .93 & .86 & \textbf{.89} \\
        deathplace     & .15 & .49 & .23          & .25 & .86 & \textbf{.47} \\
        educated & .88 & .93 & \textbf{.91} & .93 & .95 & .87          \\
        occupation & .93 & .82 & .87          & .93 & .85 & \textbf{.91} \\
        parent  & .61 & .66 & .62          & .61 & .85 & \textbf{.71} \\
        sibling & .66 & .70 & .67          & .68 & .75 & \textbf{.73} \\
        child   & .59 & .58 & .58          & .72 & .58 & \textbf{.65} \\ 
        other      & .70 & .66 & \textbf{.68} & .78 & .60 & .65          \\ \hline
        \textbf{macro}      & .70 & .73 & .70          & .72 & .81 & \textbf{.74} \\ \hline
    \end{tabular}
\caption{XLM-R and mBERT trained on English dataset \protect \cite{plum2022}, evaluated on German gold dataset under cross-lingual learning.}
\label{tab:zero_learn}
\end{table}

According to the results, XLM-R outperforms the mBERT model in overall macro and weighted F1 scores. However, the important finding is that the results of the cross-lingual experiments are close to the monolingual results in Table \ref{tab:monolingual_learn}. This highlights that the transformer models trained with datasets created using GDS are capable of performing cross-lingual transfer learning. Therefore, if external resources such as Wikidata or Pantheon would not exist or be of poor quality, due to the target language being low-resource, cross-lingual transfer learning could be a viable alternative method. This finding can be beneficial for a multitude of low-resource languages, although it depends on how well cross-lingual transfer learning works for different language pairs. 

\subsection{Multilingual Learning}
Because of the promising results of cross-lingual learning, we also assessed the performance of using a combined dataset of both English and German. The aim of this approach would be to boost performance, with the model learning from both datasets, while still only targeting one language for evaluation. Overall, we do see a significant increase in both macro and weighted F1 scores. Especially precision in quite low-performing relations, such as \textit{deathplace} and \textit{sibling}, we see an increase. In terms of models, it is clear that XLM-R benefits more from this data, as opposed to only minor improvements with mBERT. These findings show that multilingual learning is beneficial with the datasets created with GDS, highlighted by Table \ref{tab:combined_learn}.

\begin{table}[t!]
\centering
    \begin{tabular}{|l|ccc|ccc|}
        \hline
        \textit{Multilingual} & \multicolumn{3}{c|}{\textbf{mBERT}} & \multicolumn{3}{c|}{\textbf{XLM-R}} \\
        \textbf{Relation} & \textbf{P} & \textbf{R} & \textbf{F1} & \textbf{P} & \textbf{R} & \textbf{F1} \\ \hline
        birthdate         & .97 & .95 & \textbf{.96} & .95 & .97 & \textbf{.96} \\
        birthplace            & .72 & .95 & \textbf{.81} & .66 & .87 & .80          \\
        deathdate         & .93 & .96 & .94          & .94 & .96 & \textbf{.96} \\
        deathplace            & .24 & .83 & .35          & .34 & .85 & \textbf{.40} \\
        educated        & .90 & .96 & \textbf{.93} & .95 & .82 & .87          \\
        occupation        & .78 & .99 & .87          & .93 & 1.0 & \textbf{.96} \\
        parent         & .63 & .82 & .71          & .71 & .72 & \textbf{.77} \\
        sibling        & .64 & .67 & .65          & .75 & .77 & \textbf{.78} \\
        child          & .67 & .60 & .63          & .90 & .58 & \textbf{.65} \\ 
        other             & .84 & .53 & .65          & .81 & .78 & \textbf{.77} \\ \hline
        \textbf{macro}    & .72 & .84 & .75          & .77 & .80 & \textbf{.77} \\ \hline
    \end{tabular}
\caption{XLM-R and mBERT trained on English \protect \cite{plum2022} and German dataset, evaluated on German gold set.}
\label{tab:combined_learn}
\end{table}

Comparing the results of Multilingual BERT trained on the English and German multilingual dataset, we find it outperforms the same model trained in the cross-lingual setting, shown in Table \ref{tab:zero_learn}. This is most noticeable in terms of recall, while in terms of precision the difference is not as much. Compared to the results of the monolingual training, shown in Table \ref{tab:monolingual_learn}, the difference in recall is less, however the difference in precision is higher. Overall, the idea that the multilingual learning setting would combine the good results from the two separate datasets appears to hold true.


\subsection{Error Analysis}
It is clear that the evaluation results often show only minor differences for the models in each setting. In addition, we want to compare the best-performing models from each setting. To achieve this, we conducted an error analysis, looking at which relations each model made mistakes, and comparing where models either made similar incorrect predictions or cases where only one model made an incorrect prediction and the others were correct.

First, with \textbf{monolingual} learning, we found that the trends obtained from confusion matrices were extremely similar among the models, with the exception of the English BERT model, which made fewer incorrect predictions overall. It should be noted that the most incorrect predictions were made in the \textit{other} relation across all models. Main differences, where one model made an incorrect prediction, and others did not occur with GBERT, which distributed its incorrect predictions more evenly across all relations than mBERT, with XLM-R being mainly concentrated in the \textit{other} relations.

Second, in the \textbf{cross-lingual} setting, we found the trends obtained from confusion matrices again to be extremely similar. In terms of incorrect predictions for only one model, the XLM-R model only made these in \textit{other} and \textit{birthplace}, while mBERT distributed these across all relations.

Third, for the \textbf{multilingual} setting, trends were found to be similar again. The XLM-R model shows much better performance in the \textit{other} relations. In terms of single model divergence per relation, we found the same trend as in the \textit{cross-lingual} setting, where XLM-R was concentrated in \textit{other} and \textit{birthplace}, with mBERT distributed across all relations.

Finally, we compared the overall best-performing approaches: the MT baseline approach using BERT, the multilingual XLM-R model, and the cross-lingual XLM-R model. We found that although the two XLM-R models perform quite closely, the MT baseline approach is better at predicting the \textit{other} relation than the multilingual XLM-R model (which performs best overall). Taking the divergences per relation per model into account, we found that both XLM-R models produced incorrect predictions in the \textit{other} relation most, while the MT baseline was spread across all relations.


Overall, it is clear that each model produces predictions that vary across the relations, meaning that certain areas of application may favour specific models, according to the needs of that particular application. However, the quality of the dataset and, more specifically, the quality of the automatic annotation of each relation plays an important role. A relation that is high in number and encompasses many aspects of some of the other relations used here, such as the \textit{other} class, will cause classification errors, as will a relation with only a few and often ambiguous meanings, such as \textit{deathdate}. The relation \textit{deathdate} is so much fewer in number in comparison to \textit{birthdate}, as many persons did either not have a date of death because they are still alive, or it is mentioned much further towards the end of an article, making it more likely that something else (incorrect) is matched beforehand. In addition, relations that are closely related, such as the \textit{parent, sibling} and \textit{child} relations, are difficult to annotate with GDS and, therefore cause incorrect predictions with the models. Incidentally, it was also difficult for human annotators with these closely related relations, as sentences were annotated without context information from previous sentences that would have occurred in the article. 

\section{Discussion}
\label{sec:discussion}
NER and RE are two central tasks for the extraction and modelling of semantic information from text data. Combining them allows the reconstruction of basic semantic structures encoded in sentences, e.g., biographical information about persons, that humans have no problem understanding but is impossible to learn for language models without structured input. Like most other tasks in NLP, NER and RE work best for high-resource languages with many language technological resources available for model training, e.g., manually annotated datasets of semantic relations between entities. 

Automated approaches to this problem, such as \textit{distant supervision}, provide a solution that bypasses the arduous manual annotation of training datasets. At the same time, they introduce new problems for model training, be it the assumption of inter-context equality of entity relations present in the same data instance or best candidate estimation in models with sentence attention – all of which may lead to false relation labelling and under-specific taxonomies. For these problems, GDS offers a viable solution by harnessing existing sets of manually curated, structured relation data. Compared to prior solutions, GDS makes it possible to establish fine-grained and context-specific relations between entities. Additionally, considering multilingual datasets, building a RE pipeline based on parallel (multilingual) relation taxonomies provides a domain-specific solution that also reflects the semantic, grammatical and cultural differences that may be present in different languages.

There are various domains of application for GDS in multilingual text processing, ranging from improving MT applications over offering promising starting points for developing language resources for small languages with sparse data and the processing of multilingual historical text corpora in Digital Humanities (DH) projects to providing a base layer of semantic structure present in language for text processing in Large Language Models (LLMs):

\vspace{1mm}
\noindent \textit{\textbf{(1)}} While performance in current MT models is good for pairs of large languages with sufficient training data available, for low-resource languages, performance may be limited – if available at all. Here, datasets based on carefully curated sources of structured data can be helpful in improving overall model performance by providing a reliable source of semantic relations between entities. This includes bootstrapping solutions for small languages, e.g., from German to Luxembourgish, due to their close typological connection.

\vspace{1mm}
\noindent \textit{\textbf{(2)}} Especially for smaller languages with limited resources, many NLP applications still show limited or varied performance. This includes multilingual and language-specific tasks based on current neural network architectures like BERT \cite{rogers-etal-2020-primer} in general, but also in specialised tasks for small languages like Luxembourgish \cite{lothritz-etal-2020-evaluating, lothritz-etal-2022-luxembert}. Having access to datasets of semantically meaningful relations in the text can be beneficial to the development of new language resources for small languages by providing additional training data for information extraction and language modelling for a variety of tasks.

\vspace{1mm}
\noindent \textit{\textbf{(3)}} For DH projects occupied with the processing of large multilingual corpora, e.g., historical newspapers \cite{duering-etal-2021-impresso-inspect}, tasks like NER and RE are crucial to developing datasets and platforms that provide access to sources that go beyond the statistical aggregation of form occurrences in text data, i.e., by adding a semantic layer of structured entity relations that is available in different languages and thus improves document search and network graph modelling.

\vspace{1mm}
\noindent \textit{\textbf{(4)}} More generally speaking, and in light of the recent discussion around performance and reliability in LLMs, our approach might come in handy when considering the state and further development of language models. One main criticism in this regard is the fact that even state-of-the-art models that output convincingly-looking text, e.g., recent conversational agents like ChatGPT\footnote{\url{https://openai.com/blog/chatgpt/}}, are nothing more than statistical automata for text generation predicting the most likely word to follow given a linguistic context. In doing so, they are operating exclusively on the form level of language without access to (or even "knowledge" of) semantic relations in language. This is true even despite the fact that it has been shown various times that distance similarities in vector-based language models seem to reflect certain aspects of semantic differences between concepts derived from their use in similar contexts \cite{doi:10.1073/pnas.1720347115, doi:10.1177/0003122419877135}. To move beyond this state in LLMs, it will be necessary to explore new ways of adding carefully curated datasets mirroring semantic aspects of language use, for which our approach offers a promising starting point.

\section{Conclusion}\label{sec:conclusion}
In this paper, we have presented a dataset for German biographical relation extraction, compiled using GDS. We have described the steps taken to adapt the already existing methodology for a new language. Additionally, a thorough investigation has been carried out to assess the performance of such a dataset for relation extraction models. We also compared the performance to the results of cross-lingual learning and combining multiple datasets. Not only has this demonstrated that GDS is a valid approach for compiling data in multiple languages, but also that in low-resource settings, cross-lingual learning can perform just as well.

In the future, we would like to address some problems arising from the conversion of the methodology from English to different low resource languages. We would be exploring how the multilingual models perform in a low-resource environment for RE. Furthermore, we would like to apply the dataset and trained models to ``real-world'' data.

\section{Bibliographical References}\label{sec:reference}

\bibliographystyle{lrec-coling2024-natbib}
\bibliography{anthology,custom}


\end{document}